\documentclass[letterpaper]{article} % DO NOT CHANGE THIS
\usepackage{aaai25}
\usepackage{times}  % DO NOT CHANGE THIS
\usepackage{helvet}  % DO NOT CHANGE THIS
\usepackage{courier}  % DO NOT CHANGE THIS
\usepackage[hyphens]{url}  % DO NOT CHANGE THIS
\usepackage{graphicx} % DO NOT CHANGE THIS
\usepackage{natbib}  % DO NOT CHANGE THIS AND DO NOT ADD ANY OPTIONS TO IT
\usepackage{caption} % DO NOT CHANGE THIS AND DO NOT ADD ANY OPTIONS TO IT
\usepackage{algorithm}
\usepackage[noend]{algorithmic}

\usepackage{newfloat}
\usepackage{listings}
\usepackage{amsmath,amsfonts,amssymb,amsthm,cancel,siunitx,
calculator,calc,mathtools,empheq,latexsym}
\usepackage{multirow}
\usepackage{xcolor}
\theoremstyle{plain}
\newtheorem{assumption}{Assumption}
\newcommand{\squishlist}[1][$\bullet$]
{
    \begin{list}{#1}
    { 
        \setlength{\topsep}{-0.5em}
        \setlength{\leftmargin}{1em}
        \setlength{\labelwidth}{1em}
        \setlength{\labelsep}{0.5em}
    } 
}

\newcommand{\squishend}{\end{list}}

\newcommand{\squishenum}
{
    \begin{itemize}
    { 
        \setlength{\leftmargin}{1em}
        \setlength{\labelwidth}{1em}
        \setlength{\labelsep}{0.5em}
    } 
}
\DeclareCaptionStyle{ruled}{labelfont=normalfont,labelsep=colon,strut=off} % DO NOT CHANGE THIS
\floatstyle{ruled}
\newfloat{listing}{tb}{lst}{}
\floatname{listing}{Listing}
\title{Optimizing Ride-Pooling Operations with Extended Pickup and Drop-Off Flexibility}
\newcommand{\METHODX}{FlexiPool}
\author{
    Hao Jiang%\textsuperscript{\rm 1}
    ,
    Yixin Xu%\textsuperscript{\rm 2}
    ,
    Pradeep VARAKANTHAM%\textsuperscript{\rm 1}
}
\affiliations{
    Singapore Management University\\
    haojiang.2021@phdcs.smu.edu.sg, 
    yixinxu@smu.edu.sg, pradeepv@smu.edu.sg
}

\usepackage{bibentry}
\begin{document}

\maketitle

\begin{abstract}

The Ride-Pool Matching Problem (RMP) is central to on-demand ride-pooling services, where vehicles must be matched with multiple requests while adhering to service constraints such as pickup delays, detour limits, and vehicle capacity. Most existing RMP solutions assume passengers are picked up and dropped off at their original locations, neglecting the potential for passengers to walk to nearby spots to meet vehicles. This assumption restricts the optimization potential in ride-pooling operations.

In this paper, we propose a novel matching method that incorporates extended pickup and drop-off areas for passengers. We first design a tree-based approach to efficiently generate feasible matches between passengers and vehicles. Next, we optimize vehicle routes to cover all designated pickup and drop-off locations while minimizing total travel distance. Finally, we employ dynamic assignment strategies to achieve optimal matching outcomes.

Experiments on city-scale taxi datasets demonstrate that our method improves the number of served requests by up to 13\% and average travel distance by up to 21\%
compared to leading existing solutions, underscoring the potential of leveraging passenger mobility to significantly enhance ride-pooling service efficiency.
\end{abstract}

\section{Introduction}
The rise of on-demand transportation services has transformed urban mobility by offering users convenient, flexible travel options. One key innovation is on-demand ride-pooling (e.g., UberPool, LyftLine, GrabShare), which allows multiple passengers traveling along similar routes to share a ride. This approach not only optimizes vehicle utilization, reducing the environmental impact and traffic congestion, but also lowers costs for passengers and increases earnings for drivers, creating a more sustainable model for urban transportation.

A central problem in ride-pooling is effectively matching multiple passenger requests to individual vehicles, typically represented using the Ride-Pool Matching Problem (RMP)~\cite{alonso2017demand,shah2020neural}. The RMP aims to assign passenger requests to vehicles in a way that adheres to specific service constraints, such as pickup delays, detour limits, and vehicle capacity, while also optimizing an objective like maximizing the revenue or number of served requests or minimizing travel distances. This problem grows complex in city-scale applications where thousands of locations and vehicles of varying capacities are involved, requiring efficient, scalable solutions.

Traditional approaches to RMP assume fixed pickup and drop-off points~\cite{alonso2017demand,shah2020neural,hao2022hierarchical}, which limits the flexibility and efficiency of ride-pooling services. Inspired by previous research~\cite{saisubramanian2015risk} which demonstrated that flexible vehicle assignments can enhance response efficiency under uncertainty but were restricted to fixed locations.
%In this paper
, we explore a more flexible scenario where passengers can walk to nearby pickup and drop-off points. This flexibility not only reduces waiting times but also improves match success rates and minimizes unnecessary detours, ultimately enhancing system performance. However, allowing flexible pickup and drop-off locations significantly increases the potential combinations and complicates the matching process.

Recent research~\cite{gao2024lightweight} has started to explore flexible pickup and drop-off options; however, it primarily focuses on short-term optimizations without considering the long-term effects on vehicle distribution and future matches. 
%{\color{red} Furthermore, they make the unrealistic assumption of  pick up or drop off from any point in the map and not just from designated points}. 
Our approach, named \METHODX, addresses these limitations by using reinforcement learning to optimize ride-pooling matches while considering both immediate and long-term outcomes. By accounting for how each match affects future vehicle availability and routing options, our method improves the overall operational efficiency of the ride-pooling system significantly.

The \METHODX\ approach consists of three main steps. First, we employ a tree-based method to efficiently generate feasible combinations of passenger requests for each vehicle, allowing us to prune infeasible matches early and reduce the search space. Second, we optimize vehicle routes to cover designated pickup and drop-off locations by formulating this as a regional route-planning problem and solving it through mixed integer linear programming. Finally, we utilize a neural approximate dynamic programming technique, HIVES~\cite{hao2022hierarchical}, to evaluate each assignment in terms of both immediate and future rewards, enabling the system to make informed, forward-looking decisions.

To validate our approach, we conducted experiments on a large, real-world taxi dataset~\cite{yellowtaxi}. Results demonstrate that \METHODX\ outperforms leading RMP solvers~\cite{shah2020neural, hao2022hierarchical} by achieving up to a 13\% improvement in the number of served requests and a 21\% reduction in average travel distance. This improvement is significant for on-demand transportation services, where even a 0.5\% increase  in efficiency translates into notable gains in operational performance~\cite{zhou2020multi}.

%In summary, our work presents the first comprehensive approach to RMP that incorporates flexible pickup and drop-off points alongside long-term impact considerations. This method optimizes vehicle routes and improves ride-pooling efficiency, supporting a more effective and sustainable urban mobility solution.

In summary, our work makes the following contributions: 
\begin{itemize}
    \item We incorporates flexible pickup and drop-off locations for RMP, which not only increases the number of feasible matches but also reduces unnecessary vehicle detours, leading to improved overall performance.
    % This innovation reflects the conditions of urban transportation, where passengers are often willing to walk to nearby locations to optimize their travel experience. 
    %By incorporating this flexibility, our approach not only increases the number of feasible matches but also reduces unnecessary vehicle detours, leading to improved overall performance.
    \item We implement an efficient algorithm to identify all feasible matches. Our approach comprehensively explores various  assignment possibilities while maintaining computational efficiency.
    \item We formulate the Regional Vehicle Routing Problem (RVRP) to optimize the routes for vehicles based on the generated feasible combinations, ensuring that all service constraints are satisfied while minimizing travel time.
\end{itemize}

After these steps, we employ a reinforcement learning framework to evaluate the impact of each assignment. Empirically, we show our method outperforms leading RMP solvers~\cite{shah2020neural,hao2022hierarchical} by up to 13\% in the number of served requests up to 21\% in average travel distance.

\section{Related Work}

Earlier research on RMP has employed a range of optimization and planning methods, including branch-and-price algorithms, insertion heuristics, and column generation techniques~\cite{ropke2009branch,ritzinger2016survey,parragh2008survey}. These approaches have demonstrated effectiveness in small-scale applications, but they often encounter difficulties when scaling up to city-wide scenarios involving numerous requests and vehicles.

As ride-pooling problems expanded to larger settings, researchers started to focus on strategies for city-scale applications~\cite{ma2013t,huang2013large,tong2018unified,lowalekar2019zac,alonso2017demand}. 
% Many of these approaches emphasize short-term decision-making, optimizing current matches based on immediate information, but without fully considering the longer-term impacts on future assignments. 
While these methods can effectively handle large datasets, their focus on short-term optimization limits their ability to adapt to future matching needs, resulting in myopic performance over the long term.

To consider the long-term impacts of current assignments in city-scale RMPs, reinforcement learning-based (RL-based) methods such as Neural Approximate Dynamic Programming (NeurADP) are proposed~\cite{shah2020neural,hao2022hierarchical}. 
% While NeurADP predicts long-term rewards for each vehicle and offers a state-of-the-art solution, its value decomposition framework does not fully account for interactions between agents, which restricts its optimization potential in multi-agent settings.
% Route optimization is often neglected in RL-based assignment methods for the Ride-Pool Matching Problem (RMP). 
% Although RL-based methods can efficiently assign requests to vehicles in dynamic environments while considering long-term effects, they often overlook the route optimization of vehicles, which is critical and significantly affects travel times and overall system performance, especially in large-scale problems.
Although existing RL-based methods can efficiently assign requests to vehicles in dynamic environments while considering long-term effects, they do not account for flexible pickup and drop-off points and overlook the route optimization of vehicles. 

Recent research~\cite{dessouky2023ridesharing,dessouky2024stochastic} has demonstrated the advantages of incorporating flexible pickup and drop-off points in ride-pooling systems. Allowing passengers to walk a short distance to nearby meeting points can significantly reduce vehicle detours and improve travel efficiency. For instance, research indicates that flexible pickup and drop-off locations can lead to up to an 18\% reduction in travel times, as well as a significant decrease in passenger waiting and in-vehicle time~\cite{dessouky2023ridesharing}. This approach is especially effective in large-scale urban areas, where even slight adjustments to pickup and drop-off locations can bring out great optimization in vehicle utilization. Furthermore, introducing such flexibility has been shown to improve the scalability of ride-pooling systems, making them better suited for large and dynamic urban settings~\cite{dessouky2024stochastic}. However, these methods overlook the long-term rewards of current assignments, whereas our approach considers this aspect using reinforcement learning in dynamic, multi-vehicle scenarios. A recent work utilizes Graph Convolution Network (GCN) to compute optimal routes with flexible pickup and drop-off points~\cite{gao2024lightweight}. Their work uses a supervised model and requires pre-training with real-world network data, which may limit its applicability in different settings. In contrast, our method is independent of specific datasets and can be applied across various contexts.

An important issue in ride-pooling is planning vehicle routes to serve all passengers while adhering to their service constraints. With flexible pickup and drop-off points, a vehicle must select one point from a set of  designated pickup/drop-off areas while adhering to the service constraints. This problem can be formulated as the Regional Vehicle Routing Problem (RVRP), a variant of the Vehicle Routing Problem (VRP).
Modern methods have leveraged machine learning models, such as graph neural networks, to enhance the solution quality of the VRP~\cite{lin2024cross, dornemann2023solving}. However, these traditional VRP approaches are only applicable with fixed pickup and drop-off points, making them unsuitable for our work. We develop a Regional Vehicle Routing Problem (RVRP) framework that supports flexible pickup and drop-off points while accounting for the specific service constraints of the ride-pooling problem (RMP).

\begin{figure*}[htbp]
    \centering
    \begin{minipage}{0.33\textwidth}
        \centering
        \caption*{I. Collect Request from Road Network}
        \includegraphics[width=\columnwidth,height=100pt]{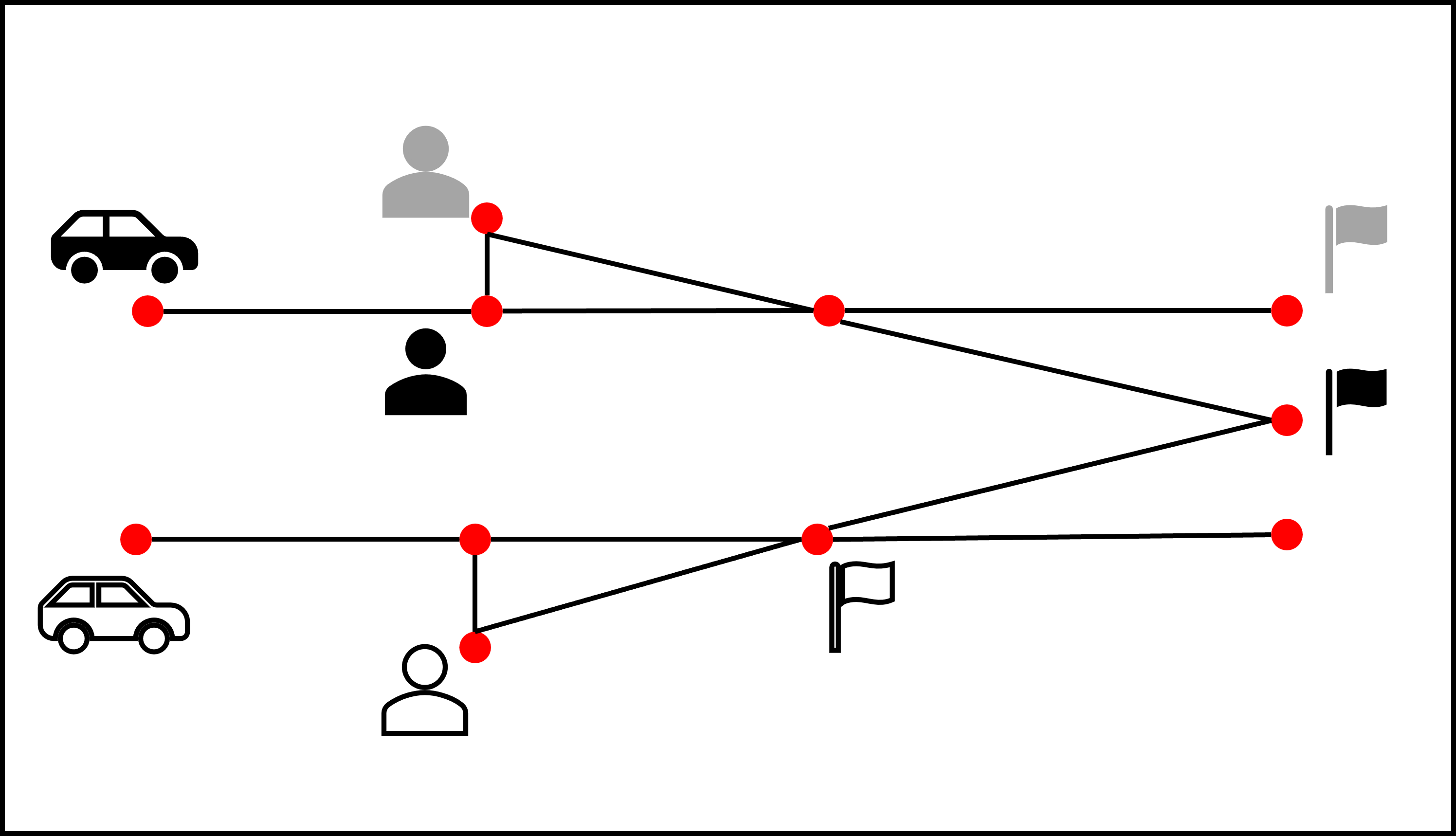}
        \label{StepA}
    \end{minipage}\hfill
    \begin{minipage}{0.33\textwidth}
        \centering
        \caption*{II. Extend Pickup/Drop-off Area}
        \includegraphics[width=\columnwidth,height=100pt]{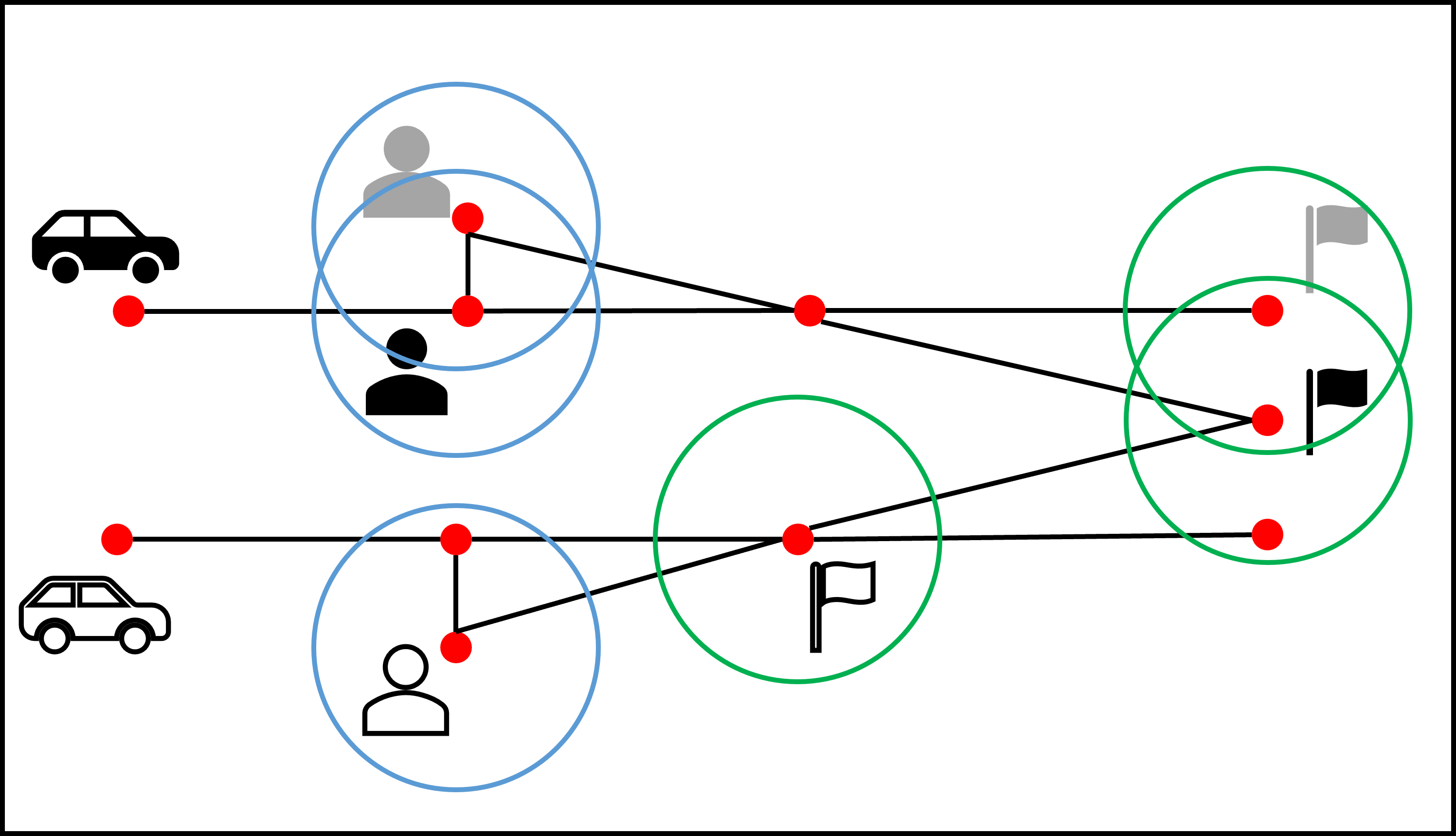}
        \label{StepB}
    \end{minipage}\hfill
    \begin{minipage}{0.33\textwidth}
        \centering
        \caption*{III. Generate Feasible Actions}
        \includegraphics[width=\columnwidth,height=100pt]{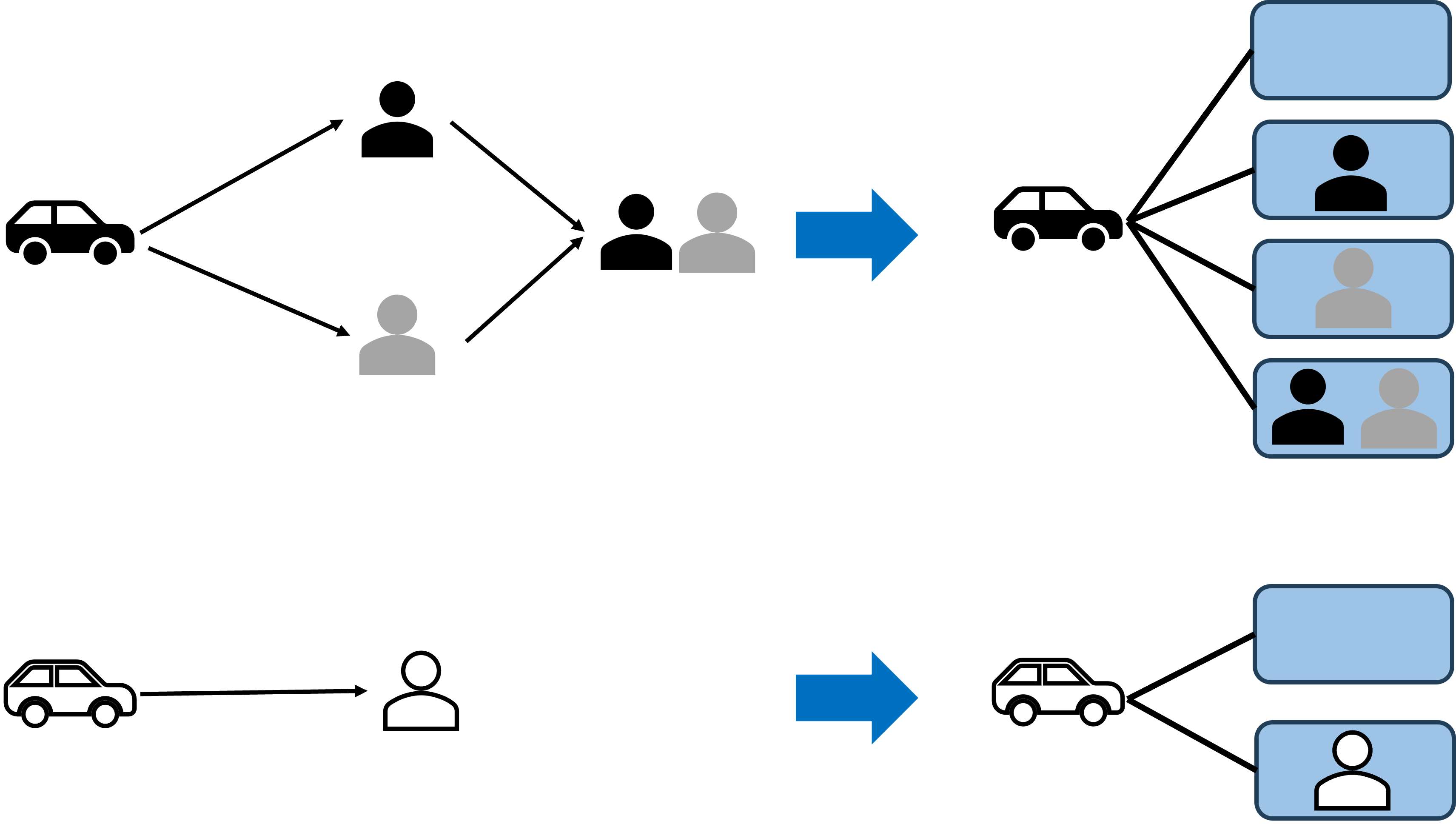}
        \label{StepC}
    \end{minipage}
    %\Description{The figure illustrates the step from receving requests from customers to score feasible actions according to the requests}
    \begin{minipage}{0.33\textwidth}
        \centering
        \caption*{IV. Score Feasible Actions}
        \includegraphics[width=\columnwidth,height=100pt]{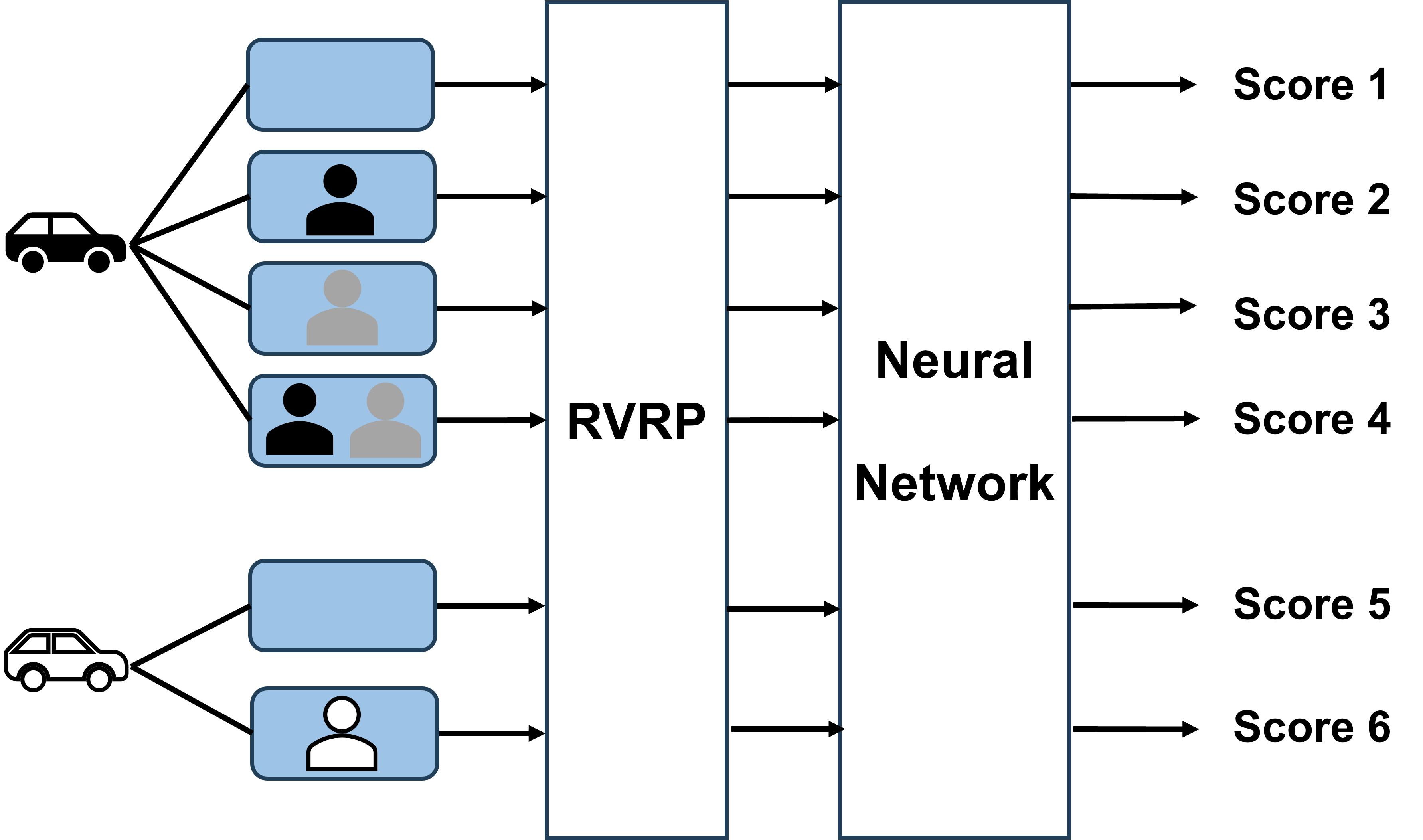}
        \label{StepD}
    \end{minipage}\hfill
    \begin{minipage}{0.33\textwidth}
        \centering
        \caption*{V. Value Function Update}
        \includegraphics[width=\columnwidth,height=100pt]{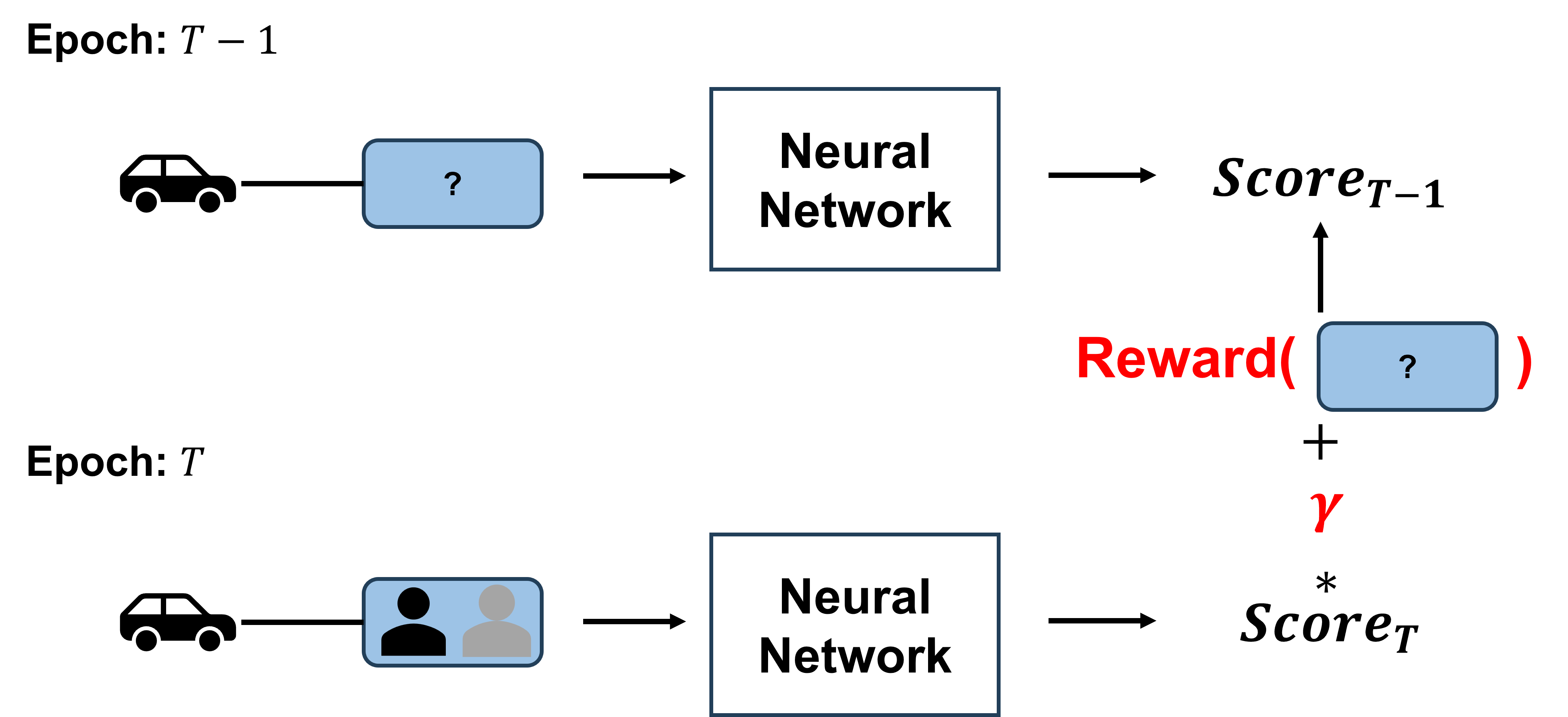}
        \label{StepE}
    \end{minipage}\hfill
    \begin{minipage}{0.33\textwidth}
        \centering
        \caption*{VI. Simulate Motion}
        \includegraphics[width=\columnwidth,height=100pt]{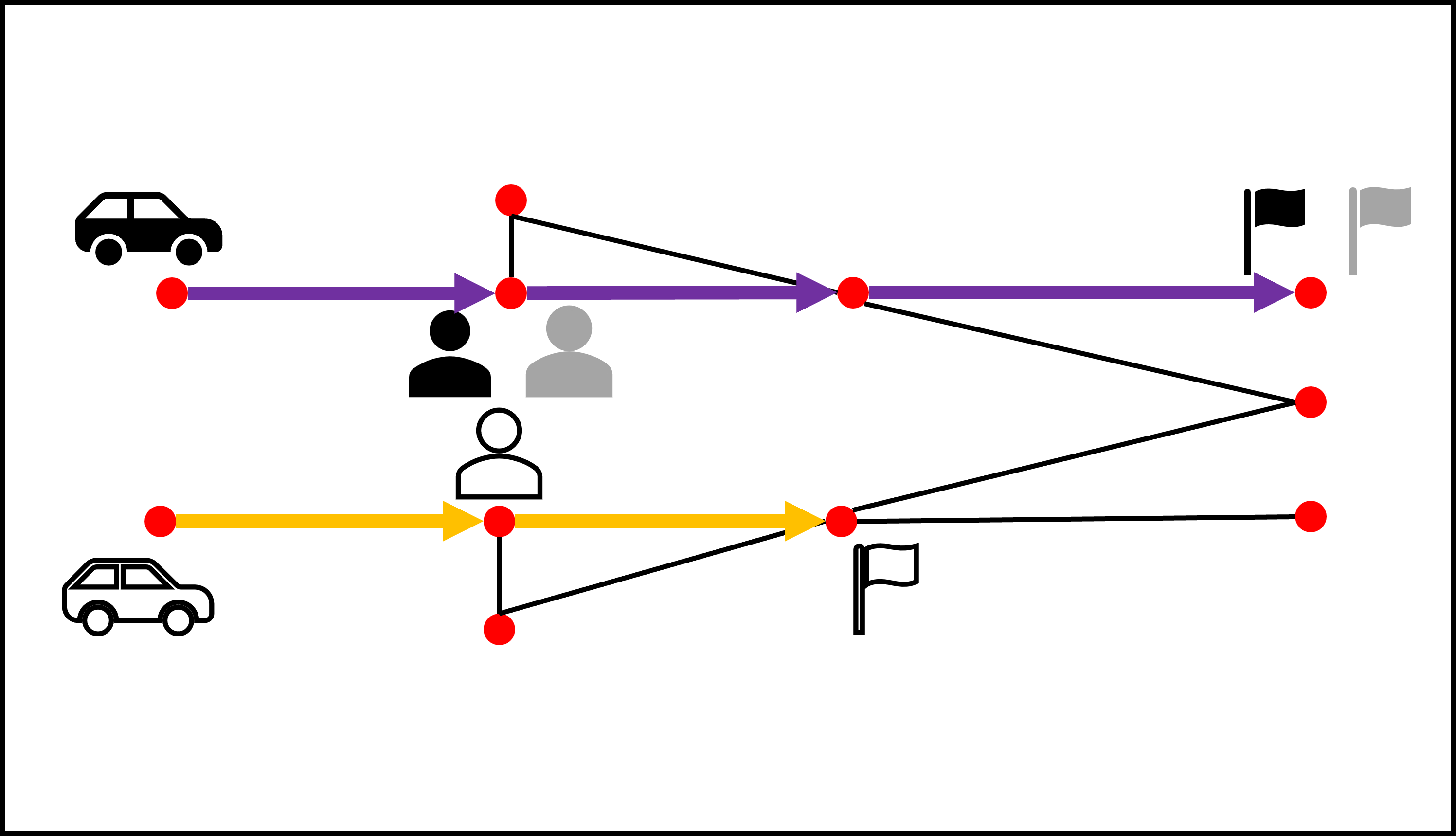}
        \label{StepF}
    \end{minipage}
    \caption{Overview of \METHODX.}
    %\Description{The figure illustrates how system does assignment and optimization based on the scores given by neural network}
    \label{fig:NeurADP}
\end{figure*}

\section{Ride-pool Matching Problem(RMP)}
Formally, traditional Ride-pool Matching Problem(RMP) can be formulated as the tuple: $\left <\mathcal{G,R,V,D},\Delta,\mathcal{O}\right > $
\squishlist
\item[$\mathcal{G}$]: $\mathcal{G} = \langle I,E \rangle$ denotes the underlying graph of the road network. $I$ are the vertices in the graph and roads connecting the intersections, $E$ are the edges. Without loss of generality, pickups and drop-offs are assumed to occur at intersections.
\item[$\mathcal{R}$]: denotes the set of user requests. $\mathcal{R}_t$ is the set of all the requests collected in epoch $t$, where $\mathcal{R}=\cup_t \mathcal{R}_{t}$. Each request ${r}_{j}$ in $\mathcal{R}$ is represented using the tuple $\left <{p}_{j},{e}_{j},t_{j} \right>$, where ${p}_{j}$ denotes the pickup location, ${e}_{j}$ denotes the drop-off location, $t_{j}$ denotes the arrival time of the request.
\item[$\mathcal{V}$]: denotes the set of vehicles. Each vehicle ${v}_{i}$ in $\mathcal{V}$ is represented using the tuple $\left<{c}_{i},{l}_{i},\mathcal{L}_{i} \right>$, where ${c}_{i}$ denotes the capacity of the vehicle, ${l}_{i}$ denotes the current location of the vehicle, $\mathcal{L}_{i}$ denotes the list of locations to go to finish the assigned requests of the vehicle. It is null if no request is assigned to vehicle $i$.
\item[$\mathcal{D}$]: considers two delay constraints. One is the maximum allowed pickup delay $\delta$. Another is the maximum allowed detour delay $\lambda$. 
\item[$\Delta$]: denotes the length of each time window.
\item[$\mathcal{J}$]: denotes the objective. We use $\mathcal{J}_{t}^{i,f}$ to represent the value that vehicle $i$ achieves in epoch $t$ on serving feasible request combination $f$.\\
\squishend

\noindent The goal in RMP is to optimalize the overall objective, $\mathcal{J}$ over all vehicles and at all epochs.

\section{Overview of \METHODX\ }
\begin{figure}[htbp]
    \centering
    \begin{minipage}{0.23\textwidth}
        \centering
        \caption*{A. Solution without flexible pickup and drop-off}
        \includegraphics[width=\columnwidth]{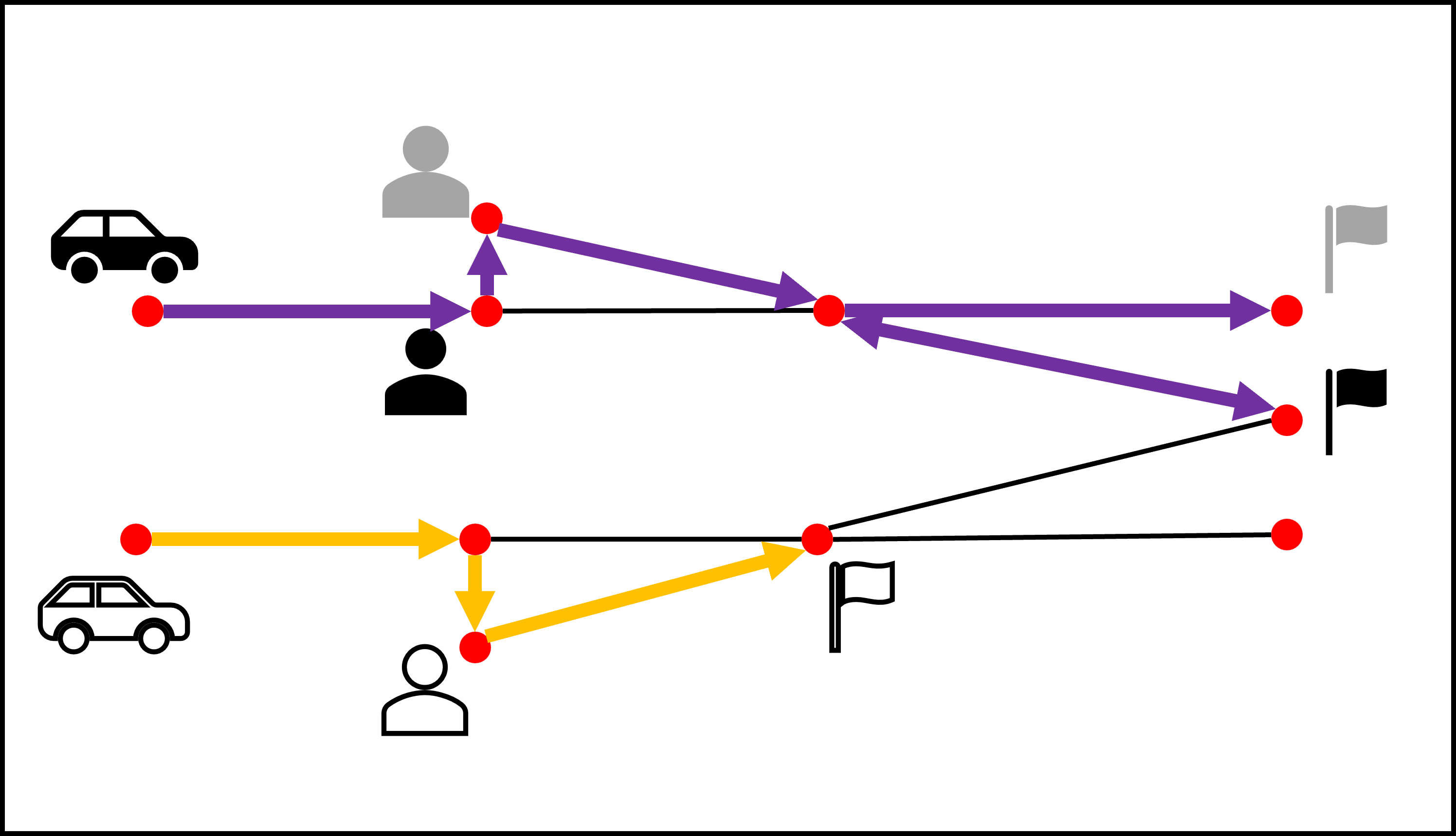}
    \end{minipage}\hfill
    \begin{minipage}{0.23\textwidth}
        \centering
        \caption*{B. Solution with flexible pickup and drop-off}
        \includegraphics[width=\columnwidth]{Step6.png}        
    \end{minipage}
    \caption{Impact of Flexible Pickup and Drop-off on Solution Quality.}
    \label{fig:2}
\end{figure}

Figure~\ref{fig:NeurADP} presents the overall flow of \METHODX. In each epoch, requests $\mathcal{R}$ and vehicles $\mathcal{V}$ are present at certain pick up points on the road network $\mathcal{G}$. In Step I, we collect these positions of customer requests and vehicle positions. For instance, there are two available vehicles (black and white) and three passengers (black, gray and white), each with specific pickup and drop-off points. In Step II, we compute the extended pickup and drop-off areas for each passenger. In Step III, we apply Algorithm \ref{alg:dcop} to generate all the feasible assignments. For example, the black taxi can either pick up the black passenger or not pick up anyone, while the white taxi can pick up either passenger or choose not to pick up anyone. In step IV, 
We compute an optimal route for each feasible assignment by formulating the problem as the RVRP. We then treat each assignment as an action in reinforcement learning and evaluate their scores based on current state using neural networks.
In Step V, we solve an Integer Linear Program (ILP) using the score from Step IV to figure out the optimal policy, the best action is then used to update the neural networks. 
Figure \ref{fig:NeurADP} shows the update process for the black taxi. In Step VI, the agents execute the optimal assignments derived from Step V by moving to the designated locations. In Figure \ref{fig:2}, we provide an example to illustrate the advantage due to flexible pickup and drop off. In the following subsections, we describe in detail the key steps of \METHODX.

\subsection{Extension of Pickup/Drop-off Area}
%\subsection{Preliminary}
In real-world ride-pooling scenarios, passengers may have multiple potential pickup and drop-off points rather than fixed locations. Passengers can walk to nearby pickup points to meet their assigned vehicles and similarly walk from nearby drop-off points after their rides. Formally, for a request $r=\left <{p},{e}\right>$, we use $P=\{p'_1,p'_2,...,p'_n\}$ to denote all possible pickup points and $E=\{e'_1,e'_2,...,e'_n\}$ to represent the set of possible drop-off points.

% %For each request ${r}_{t}^{j}=\left <{p}_{t}^{j},{e}_{t}^{j},t \right>$, we assume passenger can move to other feasible pickup or drop-off locations (${p'}_{t}^{j}$ or ${e'}_{t}^{j}$) for better request assignment policy. 
% In epoch $t$, for each request $r=\left <{p},{e}\right>$, we assume passenger can move to other feasible pickup and drop-off locations, denoted as ${p'}$ and ${e'}$, in order to achieve better request assignment policy. We define the extended pickup area as $P=\{p'_1,p'_2,...,p'_n\}$ and the extended drop-off area as $E=\{e'_1,e'_2,...,e'_n\}$. We first focus on the extension of pickup area where all the possible pickup location $p'\in P$ must satisfy assumption \ref{as:as1}.

\begin{figure}[htbp]
    \centering
    \includegraphics[scale=0.26]{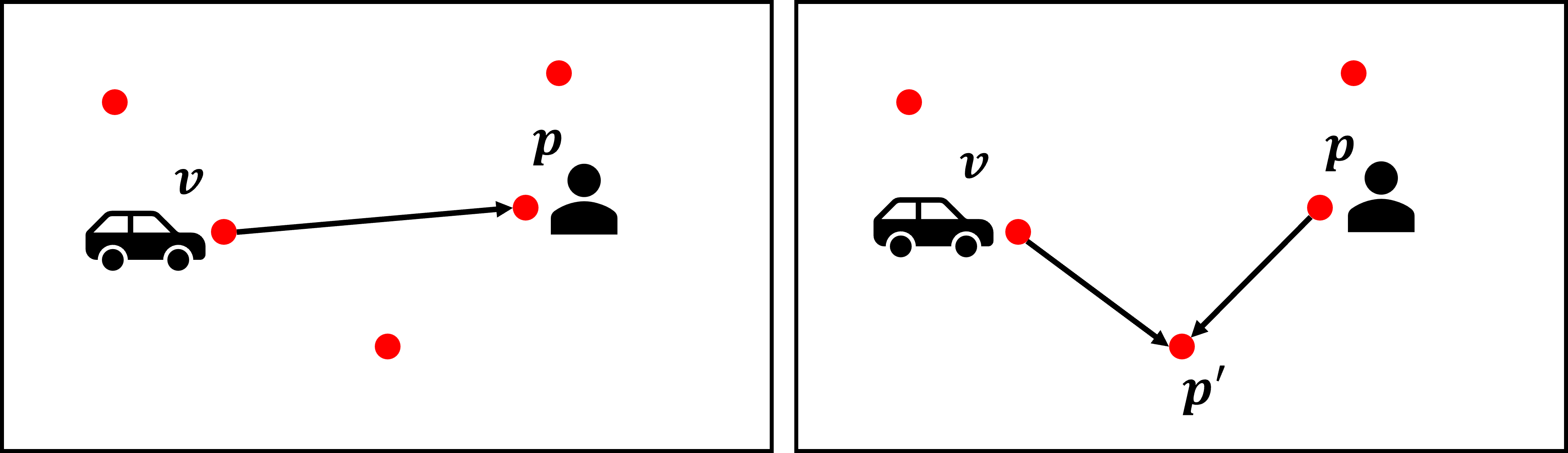}
    \caption{In the original setting (left), vehicle $v$ must travel to the passenger's original pickup location $p$. With flexible pickup and drop-off (right), passengers can walk to an alternative pickup point $p'$ that is more accessible for the vehicle. }
    \label{fig:1}
\end{figure}

% Figure \ref{fig:1} (left) illustrates the original setting where a vehicle $v$ must travel to passenger's original pickup location $p$. According to the constraint, the travel time between $v$ and $p$ must be smaller than the maximum allowed pickup delay, i.e., $\delta$: $t_v(v.p)\leq\delta$. This indicates the maximum allowed time for pickup is $\delta$. 

\begin{assumption}
\label{as:as1}
The walking time from the passenger's current location $p$ to a feasible pickup point $p'$ must be no longer than the passenger's pickup delay $\delta$.
\end{assumption}

% \textbf{Proof:} 
% As shown in Figure \ref{fig:1} (right), if a passenger's walking time exceeds their pickup delay, they cannot be picked up on time and thus violates the pickup delay constraint.

\begin{assumption}
\label{as:as2}
The driving time from a vehicle's current location $l^i$ to the pickup point $p'$ must be no longer than the passenger's pickup delay $\delta$.
\end{assumption}

For example, as shown in figure~\ref{fig:2}, assume the passenger’s maximum pickup delay is 5 mins. The passenger can only walk to locations within a 5-minute walking time. Any location requiring more than 5 minutes walking time cannot serve as a potential pickup point, as it would violate the pickup delay constraint. Similarly, a vehicle must arrive at the designated pickup location within 5 minutes to ensure it can pick up the passenger in time.

Within the maximum delay time, each passenger has a maximum allowed walking distance $d_r=\delta * walking\_speed$. A location can be considered a potential pickup point only if it is within the maximum allowed walking distance.
We define the pickup area P as: 
$$d(p,p'_i)\leq d_r, \forall p'_i\in P$$
\noindent where $d(p,p'_i)$ denotes the distance between the original pickup location $p$ and any feasible location $p'_i$ in the pickup area $P$.

%Let's use ${speed}_r$ to represent the passenger's walking speed. The travel distance between the passenger's current location to a potential pickup point must be no longer than $d_r = \delta *{speed}_r$.

% any possible pickup location $p'$ must satisfy: $\max\{t_v(v,p'),t_r(p,p')\}\leq\delta$, which means that both $t_v(v,p')\leq\delta$ and $t_r(p,p')\leq\delta$ must hold true.

% We further draw the extended pickup area $P$ for passenger $r$ with original pickup location $p$. Note that there exists the maximum allowed travel distance $d_r = t_r *{speed}_r$ for passenger, where ${speed}_r$ is a constant representing the passenger's walking speed. According to Assumption \ref{as:as1}, the threshold for $t_r$ is the maximum allowed pickup delay $\delta$. Therefore, we set maximum allowed travel distance for passenger ($d_r$) as parameter, which are positively correlated with $\delta$. 

% An example of the extended pickup area is shown in Figure \ref{fig:2}. In the original setting, vehicle $v$ needs to transport the passenger from pickup location $p$ to destination $e$. After extending the pickup area, as shown on the right, passenger can walk from $p$ to $p'$ within the feasible pickup area, thus shortening the vehicle's travel distance.

The parameter $d_r$ is also used to extend the drop-off area and generate potential drop-off points. A point is considered a potential drop-off point only if it is within $d_r$ walking distance from the original drop-off location.
% but $d_r$ must be further tuned to avoid overlapping between the extended pickup and drop-off areas for each passenger. Otherwise, the passenger could potentially walk directly to their destination, making the vehicle unnecessary.

%\begin{figure}[htbp]
%    \centering
%    \includegraphics[scale=0.26]{Picture2.png}
%    \caption{The left figure shows the path where vehicle $v$ picks up passenger at original pickup point $p$ and drops off at drop-off point $e$. The right figure presents the new setting, where passenger can walk to the optional pickup point $p'$ within the maximum walking distance $d_r$, thus shortening the vehicle’s path}
%    \textcolor{red}{(Can this figure be merged with Figure~\ref{fig:2}, they seem to convey similar information)}
%    \label{fig:2}
%\end{figure}

\subsection{Feasible Combination Generation}
In this section, we present an efficient method to generate all feasible assignments. The goal is to find all feasible assignments that match a set of requests $R=\{r_1,r_2,...,r_n\}$ to a set of vehicles $V=\{v_1,v_2,...,v_m\}$ while adhering to all constraints. By generating all feasible assignments, we can ensure that the matching process accounts for all possible options and selects the most promising ones. We propose an optimized assignment algorithm that aims to minimize unnecessary expansions in the search space, thereby reducing computational complexity. The algorithm considers the assignment for each individual vehicle and is outlined in Algorithm \ref{alg:dcop}.

\begin{algorithm}[t]
\caption{Recursive Tree Generation with Feasibility Check}
\label{alg:dcop}
\begin{algorithmic}[1]
\STATE \textbf{Input:} Vehicle capacity $c$, request set $R$, pickup delay $\delta$, detour delay $\lambda$
\STATE \textbf{Output:} Tree with feasible combinations

\STATE Initialize $Feasible\_Comb \gets \emptyset$ \COMMENT{Store feasible combinations}

\STATE Call \textbf{GenerateTree}($\emptyset$, $R$) \COMMENT{Start with empty combination and all requests}
\end{algorithmic}
\end{algorithm}

\begin{algorithm}[t]
\caption{GenerateTree($current\_comb$, $Remain\_Req$)}
\label{alg:generate_tree}
\begin{algorithmic}[1]
\STATE \textbf{Input:} Current combination $current\_comb$, Remaining requests $Remain\_Req$
\STATE \textbf{Output:} Updates $Feasible\_Comb$ with feasible combinations

\IF {\textbf{Feasibility\_Check}$(current\_comb) = \textbf{False}$}
    \STATE \textbf{Stop} expanding this branch and \textbf{go back} to the previous combination to explore other possibilities
\ENDIF

\IF {Length of $current\_comb \leq c$ and $Remain\_Req = \emptyset$}
    \STATE Add $current\_comb$ to $Feasible\_Comb$
\ENDIF

\FOR {each request $r$ in $Remain\_Req$}
    \STATE $new\_comb \gets current\_comb + r$
    \STATE $new\_Remain\_Req \gets Remain\_Req$ - $r$
    \STATE Call \textbf{GenerateTree}($new\_comb$, $new\_Remain\_Req$)
\ENDFOR

\end{algorithmic}
\end{algorithm}

For the function Feasibility\_Check(combination), a combination is considered feasible for vehicle $v_i$, if there is at least one valid path from the vehicle’s current location that enables it to complete all assigned requests without violating the specified pickup delay $\delta$ and detour delay $\lambda$ constraints for each request. If such a path is found, the combination is considered as feasible, and the function proceeds to the next combination without further checking other paths. If no feasible path exists, the combination is marked as infeasible.
Any new combination that includes this infeasible set will also be infeasible. For example, if $[r_1,r_2]$ is an infeasible match for $v_1$, then any combination that contains $[r_1, r_2]$, such as $[r_1,r_2,r_3]$ or $[r_1, r_2,r_4,r_5]$, will also be infeasible for $v_1$. Therefore, once a combination is identified as infeasible, it can be eliminated from further consideration, making the search for feasible matches more efficient.

Based on this observation, we propose to generate combinations incrementally using a tree structure. We start by considering single requests (First Level), then we move on to combinations of two requests (Second Level), followed by combinations of three requests (Third Level), and so forth. 
If a combination violates any constraint—such as exceeding the vehicle's capacity or violating the pickup/detour delay—it is pruned and stored in memory to avoid considering it in further combinations at later levels.
By assessing the feasibility of smaller sets before progressing to larger combinations, our method efficiently eliminates infeasible options early on, thus significantly reducing the search space and improving overall efficiency.

% To improve efficiency, once a combination violates a given constraint—such as exceeding the vehicle's capacity or violating the pickup/detour delay—the corresponding branch is  pruned and stored in memory. This ensures that not only is the current branch pruned, but that any future combinations containing this pruned branch will be skipped. This approach significantly reduces the search space by preventing redundant explorations of invalid paths.

\begin{figure}[htbp]
    \centering
    \includegraphics[scale=0.25]{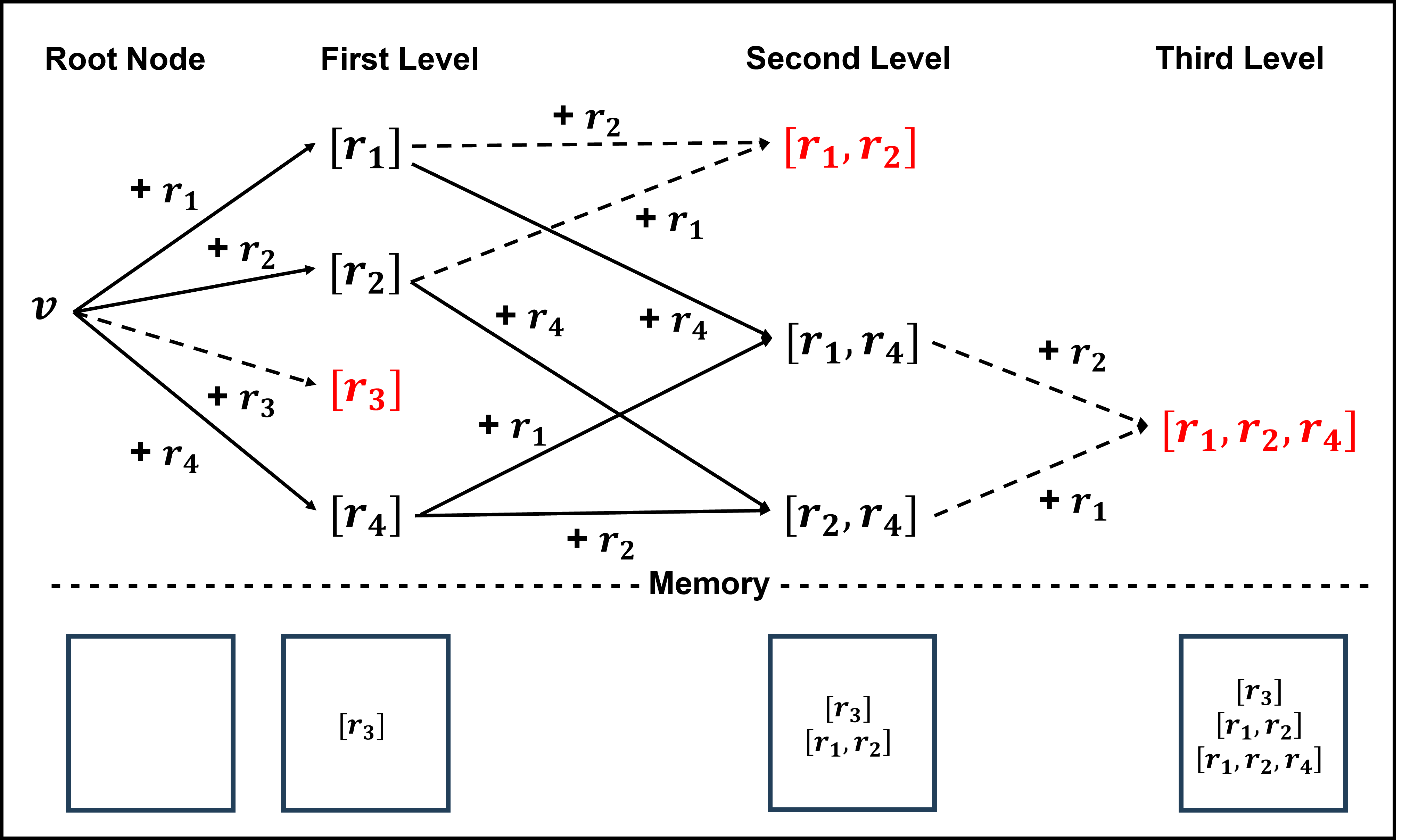}
    \caption{Example of Combination Generation. Black solid lines represent feasible combinations, while black dashed lines represent infeasible combinations.}
    \label{fig:3}
\end{figure}

Figure \ref{fig:3} illustrates the process of generating the tree using an example. Considering the matching of a vehicle $v_i$ to four requests $r_1, r_2, r_3,$ and $r_4$,
the tree begins from a root node $v_i$, indicating the vehicle is not assigned to any requests.
The first level of the tree examines the assignment of a single request. After checking the feasibility of each node when expanding the tree, any infeasible request will be pruned and stored in the memory, so that it cannot be further expanded and be considered for further combinations. For example, $r_3$ is infeasible for $v_i$, then $r_3$ will be pruned and can not be considered for further combinations. As a result, the first level will generate feasible assignments:$[r_1], [r_2], [r_4]$.

The second level of the tree examines combinations of two requests by pairing only the remaining feasible requests, specifically $[r_1], [r_2]$, and $[r_4]$. 
Initially, this level generates all combinations of two: $[r_1,r_2]$, $[r_1,r_4]$, and $[r_2,r_4]$.
If any combination is deemed infeasible, such a combination will be pruned and not considered in subsequent levels. For example, if $[r_1, r_2]$ is found to be infeasible, it will be pruned and cannot be included for further combinations.
After pruning out $[r_1, r_2]$, the second level will generate feasible assignments $[r_1,r_4]$ and $[r_2,r_4]$.

The third level of the tree examines combinations of three requests. Based on the results of the previous level, there is only one combination of three: $[r_1,r_2,r_4]$. However, this combination includes $[r_1,r_2]$, which has already been identified as infeasible and stored in memory. As a result, $[r_1,r_2,r_4]$ is added into memory 
%no feasible assignment will be generated at the third level, 
and the generation process stops here.
The final feasible assignment for $v_i$ includes the combinations generated from all levels: \{$\emptyset, [r_1], [r_2], [r_4], [r_1,r_4],[r_2,r_4]$\}. 

\subsection{Route Optimization}
After identifying all feasible matches, the next step is to plan an optimal route for each feasible match.
For a feasible match, a vehicle may be assigned to multiple requests, requiring it to first visit the pickup area and then the drop-off area for each assigned request. 
The optimal route should cover all designated areas, meet the service constraints of all assigned requests, and minimize total travel time.

We define this problem as Regional Vehicle Routing Problem (RVRP).
For each vehicle $v_i$, we assume there are $N_i$ requests awaiting pickup and $M_i$ requests already on-board.
For requests awaiting pickup, the vehicle must visit both their pickup area and their drop-off area. For requests already on board, the vehicle only needs to visit their drop-off area.
Given the capacity constraint, we have $N_i+M_i\leq c_i$. 

We formulate this problem as a Mixed-Integer Linear Program (MILP) with the following constraints and objective function:

%Firstly, we generate the feasible visiting sequence of all the areas. For the $N_i$ requests to pick up, vehicle needs to visit both pickup and drop-off areas. For the $M_i$ requests that have already picked up, only the drop-off areas need to be visited. Therefore, the visiting sequence includes $2N_i+M_i$ areas, which are derived from the sets \{$(P_1,E_1)$, $(P_2,E_2)$,...,$(P_{N_i},E_{N_i})$,$E'_1$,$E'_2$,...,$E'_{M_i}$\}.

\textbf{Decision Variables:}
\begin{itemize}
    \item $x_{ab}$: Binary variable indicating whether the vehicle travels from area $a$ to area $b$.
    \item $t_a$: Continuous variable representing the time the vehicle arrives at area $a$.
\end{itemize}

\textbf{Parameters:}
\begin{itemize}
    \item $t_{ab}$: Travel time between areas $a$ and $b$.
    \item $M$: A large constant.
    \item $v$: The starting point (initial location of the vehicle).
    \item $P = \{P_1, P_2, ..., P_{N_i}\}$: Pickup areas for requests awaiting pickup.
    \item $E = \{E_1, E_2, ..., E_{N_i}\}$: Drop-off areas for requests awaiting pickup.
    \item $E' = \{E'_1, E'_2, ..., E'_{M_i}\}$: Drop-off areas for requests already on board.
\end{itemize}

\textbf{Objective Function:}

Minimize the total travel time:
\[
\text{minimize} \quad \sum_{a \in P \cup E \cup E'} \sum_{b \in P \cup E \cup E'} t_{ab} \cdot x_{ab}
\]

\textbf{Constraints:}

1. Vehicle must start from the initial point \( v \):
\[
\sum_{b \in P \cup E \cup E'} x_{vb} = 1
\]

2. The vehicle must visit exactly one point from each pickup area \( P_j \):

For each \( P_j \), where $j \in \{1, ..., N_i\}$:
\[
\sum_{c \in P \cup E \cup E'} x_{ca} = \sum_{b \in P \cup E \cup E'} x_{ab} = 1 \quad \forall a \in P_j
\]

3. The vehicle must visit exactly one point from each drop-off area \( E_j \) and \( E'_k \):

For each \( E_j \), where $j \in \{1, ..., N_i\}$:
\[
\sum_{c \in P \cup E \cup E'} x_{cb} = \sum_{d \in P \cup E \cup E'} x_{bd} = 1 \quad \forall b \in E_j
\]

For each \( E'_k \), where $k \in \{1, ..., M_i\}$:
\[
\sum_{c \in P \cup E \cup E'} x_{cb} = \sum_{d \in P \cup E \cup E'} x_{bd} = 1 \quad \forall b \in E'_k 
\]

4. Time continuity constraint (for every transition between two areas):
\[
t_b \geq t_a + t_{ab} - M \cdot (1 - x_{ab}) \quad \forall a, b \in P \cup E \cup E'
\]

5. Pickup-Delivery sequence constraint (ensuring that pickup occurs before drop-off):
\[
t_{a} \leq t_{b} \quad \forall j \in \{1, ..., N_i\}, \text{ where } a \in P_j \text{ and } b \in E_j
\]

6. Pickup delay constraint:
\[
t_a - t_v\leq \delta \quad \forall j \in \{1, ..., N_i\}, , \text{ where } a \in P_j
\]

7. Drop-off delay constraint:
\[
t_b - t_a \leq \lambda \quad \forall j \in \{1, ..., N_i\}, \text{ where } a \in P_j \text{ and } b \in E_j
\]

8. Drop-off delay constraint for additional drop-off areas: 
\[
t_b - t^k_{start} \leq \lambda \quad \forall k \in \{1, ..., M_i\}, \text{ where } b \in E'_k
\]
where \( E'_k \) is the \( k \)-th request, \( t^k_{start} \) is the start time for the \( k \)-th request.

\subsection{Future aware Matching}
In the next step, we assess the quality of all feasible assignments considering their long-term effects. The best assignments will then be chosen and executed to optimize overall outcomes.

In this part, we apply the Neural Approximate Dynamic Programming algorithm~\cite{shah2020neural}, the leading approach for solving RMP. 
The Approximate Dynamic Programming (ADP) is commonly applied in large-scale vehicle routing and resource allocation problems. In the context of Ride-Pool Matching (RMP), ADP provides an effective method for managing dynamic vehicle assignments by predicting future rewards associated with each assignment. This prediction is based on an approximation of the system’s state and the decisions made by the agents.

Formally, The ADP problem for RMP is formulated using the tuple $\left<S,A,\xi,T,\mathcal{J}\right>$, where :
\squishlist
\item[${S}$] : denotes current state of the system, which includes the state of all vehicles $v_{t}$ and the set of available requests $r_{t}$. The state is collected in Step I of Figure \ref{fig:NeurADP}.
\item[${A}$] : denotes the set of actions, where each action corresponds to assigning a combination of requests to a vehicle. The generation of feasible combinations for each vehicle $i$ at time $t$, denoted as $\mathcal{F}^i_t$, is computed in Step III of Figure \ref{fig:NeurADP}.
%\begin{align}
%    \mathcal{F}_t^i = \{f^i | f^i \in \cup_{c'=1}^{c^i} [\mathcal{U}]^{c'}, &\text{PickUpDelay}(f^i,i) \leq \delta,\nonumber\\ 
%    &\text{DetourDelay}(f^i,i) \leq \lambda\}
%\end{align}
\item[${\xi}$]: denotes the exogenous information, which is the resource of randomness in the system. 
%This would correspond to the user requests or demand. $\xi_t$ denotes the exogenous information at time $t$. 
\item[${T}$]: denotes the transition function. In an ADP, the system evolution happens as $$(s_{0},a_{0},s_{0}^{a},\xi_{1},s_{1},a_{1},s_{1}^{a},\cdots,s_{t},a_{t},s_{t}^{a},\cdots),$$ where $s_{t}$ denotes the pre-decision state at decision epoch $t$ and $s_{t}^{a}$ denotes the post-decision state~\cite{powell2007approximate}. The transition from state $s_{t}$ to $s_{t+1}$ depends on the action vector $a_{t}$ and the exogenous information $\xi_{t+1}$. Therefore, $$s_{t+1} = {T}(s_{t},a_{t},\xi_{t+1})$$ 
$$s_{t}^{a} = {T}^{a}(s_{t},a_{t}); s_{t+1} = {T}^{\xi}(s_{t}^{a},\xi_{t+1}) $$ 
%It should be noted that ${T}^a(.,.)$ is deterministic as uncertainty is extrinsic to the system.  
\item[$\mathcal{J}$]: denotes the reward function.\\
\squishend

At each decision epoch, the goal of ADP is to find the optimal action $a_t$ by solving the following Bellman equation:
\begin{align}
 V(s_{t}) = \max_{a_{t} \in A_{t}} (\mathcal{J}(s_{t},a_{t}) + \gamma \mathbb{E}[V(s_{t+1})|s_{t},a_{t},\xi_{t+1}])
 \label{eqn:bellman}
\end{align}
where $V(s_{t})$ denotes the value function at state $s_{t}$, $\gamma$ is the discount factor. 

Using post-decision state, the Bellman equation can be reformulated into two steps:
\begin{align}
V(s_{t}) &= \max_{a_{t} \in A_{t}} (\mathcal{J}(s_{t},a_{t}) + \gamma V^{a}(s_{t}^{a})) \label{eqn:1}\\
V^{a}(s_{t}^{a}) &= \mathbb{E}[V(s_{t+1})|s_{t}^{a},\xi_{t+1}] \label{eqn:2}
\end{align}
To efficiently handle the complexity of post-decision states, we apply value decomposition. The joint value function $V^{a}(s_{t}^{a})$ is broken down into individual vehicle values:
$$V^a(s_t^a) = \sum_{i} V^{i,a}(s_{t}^{i,a})$$

\noindent where individual value function $V^{i,a}(.)$ is modeled as a neural network. This network is updated over time, using exogenous information %(e.g., observed demand from data)
and the best action determined by the ILP solution.

To incorporate the value decomposition, the following constraints for joint actions, $a_t \in \mathcal{A}_t$ is introduced: 

\begin{enumerate}
    \item Each vehicle \( i \) can only be assigned at most one request combination \( f \).
    \item At most one vehicle \( i \) can be assigned to a request \( j \).
    \item A vehicle \( i \) can either be assigned to a request combination or not.
\end{enumerate}

To mathematically formalize these constraints, let $z_{t}^{i,f}$ denote the decision variable that indicates whether vehicle $i$ takes action $f$ (a combination of requests) at decision epoch $t$. The following constraints are then applied to ensure that the above conditions are met:
\begin{align}
& \sum_{f \in {\mathcal F}_{t}^{i}} z_{t}^{i,f} = 1 ::: \forall i \in {\cal V} \label{cons:a1}\\ 
& \sum_{i \in {\mathcal V}} \sum_{f \in {\mathcal F}_{t}^{i};j \in f} z_{t}^{i,f} \leq  1 ::: \forall j \in {\cal R}_{t} \label{cons:a2}\\ 
& z_{t}^{i,f} \in \{0,1\} ::: \forall i,f  \label{cons:a3}
\end{align}

%As for the objective of ILP, the goal is to maximize the overall expected reward. The future value, $V^a(s_t^a)$ is non-linear and non-linear value functions, unlike their linear counterparts, cannot be directly integrated into the objective of ILP. One way to incorporate them is to evaluate the value function for all possible post-decision states and then add these values as constants. However, the number of post-decision states is exponential in the number of resources/vehicles. 
%
%This allows NeurADP to get around the \textit{combinatorial explosion of the post-decision state of all vehicles. } 
%Thus, the overall ILP is given by: 
%\begin{align} \label{ILP}
%\max &\sum_{i} \sum_{f \in {\mathcal F}_t^i} \left [o_t^{i,f}+V^{i,a}(T^{i,a}(s_t^{i},f))\right ] \cdot z_t^{i,f} \\
%\intertext{Subject to constraints in equations~\ref{cons:a1},~\ref{cons:a2} and \ref{cons:a3}} \nonumber
%\end{align}
%

%The individual value function $V^{i,a}(.)$ is a neural network that is updated by stepping forward through time using sample realizations of exogenous information (i.e. demand observed in data) and best action computed by the ILP. In the objective,  $V^{i,a}$ values for all possible $s_{t}^{i,a}$ (from the individual value neural network) and then integrate the overall value function into the ILP as a linear function over these individual values. This reduces the number of evaluations of the non-linear value function from exponential to linear in the number of vehicles.

Neural Approximate Dynamic Programming (NeurADP) provides a good estimate of the joint value function. HIVES \cite{hao2022hierarchical} further improves the accuracy by introducing a hierarchical mixing neural network. This network clusters agents and combines their individual value functions more effectively. 
%\textcolor{red}{So which algorithm does you apply in this work? And how is it applied?}\textcolor{blue}{
In this work, we utilize the value function structure of HIVES within our reinforcement learning framework
% Additionally, it considers the pre-decision states of neighboring agents instead of post-decision states, using $V^{i,a}(s_{t}^{i,a},\cup_{j \in N_i} s_{t}^{j})$ instead of just $V^{i,a}(s_{t}^{i,a})$, leading to better performance.

\begin{table*}[htbp]
\centering
\begin{tabular}{|c | c | c c c c c|}
\hline
 & \textbf{Variants} & NeurADP & HIVES & \METHODX & Improvement & Improvement\\
 & & & & & over NeurADP(\%) & over HIVES(\%)\\
 \hline
 \multirow{2}{*}{\textbf{Capacity}} 
 & 2 & 163465 & 181023 & 190842 & 16.75 & 5.42\\
 & 4 & 239842 & 255562 & 271882 & 13.36 & 6.39\\
 \hline
 \multirow{2}{*}{\textbf{Vehicles}}
 & 1000 & 239842 & 255562 & 271882 & 13.36 & 6.39\\
 & 1500 & 297281 & 304615 & 324446 & 9.14 & 6.51\\
 \hline
 \multirow{3}{*}{\textbf{Pickup Delay}} 
 & 300 ($d_r$=0.3 km) & 239842 & 255562 & 271882 & 13.36 & 6.39\\
 & 360 ($d_r$=0.4 km) & 240188 & 257481 & 275194 & 14.57 & 6.88\\
 & 420 ($d_r$=0.5 km) & 240753 & 260612 & 279665 & 16.16 & 7.31\\
 \hline
\end{tabular}
 \caption{Number of Served Requests Improvement over baselines.}
 \label{table:1}
\end{table*}

\begin{table*}[htbp]
\centering
\begin{tabular}{|c | c | c c c |}
\hline
 & \textbf{Variants} & NeurADP & HIVES & \METHODX\\
 \hline
 \multirow{2}{*}{\textbf{Capacity}} 
 & 2 & 345.34 & 311.84 & 285.49\\
 & 4 & 501.71 & 470.84 & 405.44\\
 \hline
 \multirow{2}{*}{\textbf{Vehicles}}
 & 1000 & 501.71 & 470.84 & 405.44\\
 & 1500 & 415.93 & 405.92 & 352.58\\
 \hline
 \multirow{3}{*}{\textbf{Pickup Delay}} 
 & 300 ($d_r$=0.3 km) & 501.71 & 470.84 & 405.44\\
 & 360 ($d_r$=0.4 km) & 502.43 & 468.69 & 402.78\\
 & 420 ($d_r$=0.5 km) & 503.64 & 465.26 & 399.52\\
 \hline
\end{tabular}
 \caption{Average Travel Distance Reduction over baselines.}
 \label{table:3}
\end{table*}

\begin{table*}[t]
\centering
\begin{tabular}{|c | c | c c c |}
\hline
 & \textbf{Variants} & Pickup Only & Drop-off Only & Pickup \& Drop-off\\
 \hline
 \multirow{2}{*}{\textbf{Capacity}} 
 & 2 & 188642 & 188755 & 190842 \\
 & 4 & 267261 & 267354 & 271882 \\
 \hline
 \multirow{2}{*}{\textbf{Vehicles}}
 & 1000 & 267261 & 267354 & 271882 \\
 & 1500 & 317056 & 318348 & 324446 \\
 \hline
 \multirow{3}{*}{\textbf{Pickup Delay}} 
 & 300 ($d_r$=0.3 km) & 267261 & 267354 & 271882 \\
 & 360 ($d_r$=0.4 km) & 270261 & 271238 & 275194 \\
 & 420 ($d_r$=0.5 km) & 274163 & 275130 & 279665 \\
 \hline
\end{tabular}
 \caption{Impact of Considering Pickup/Drop-off Location.}
 \label{table:2}
\end{table*}

\section{Experiment}
In the experiments, we aim to demonstrate the effectiveness of ride-pooling with flexible pickup and drop-off points.
We will also conduct ablation study to evaluate various configurations—extending only the pickup locations, extending only the drop-off locations, and extending both—to assess their impact on overall improvement.

\subsection{Setup}
The experiment is performed with the New York Yellow Taxi Dataset \cite{yellowtaxi}. Following the settings of similar works~\cite{shah2020neural}, we exclude locations without outgoing edges and focus on areas with the highest frequency of requests, resulting in a network consisted of 4373 locations and 9540 edges. 
We assume that all vehicles have the same capacity and are initialized at random locations.  The maximum detour delay is set to $\lambda = 2\delta$, and the time window for assignments was 60 seconds. We examine the effect of four parameters:
\begin{itemize}
    \item Vehicle capacity: varies from 2 to 4, with a default value of 4.
    \item Total number of vehicles: varies from 1000 to 2000, with a default value of 1000.
    \item Maximum pickup delay: varies from 300 seconds to 420 seconds, with a default value of 300 seconds.
    \item Maximum allowed walking distance: varies from 0.3 km to 0.5 km.
\end{itemize}
% The capacity of each vehicle was varied from 2 to 4, the total number of vehicles was varied from 1000 to 2000 and the maximum pickup delay $\delta$ was set between 300 and 420 seconds. The maximum detour delay $\lambda = 2\delta$, and the time window for assignments was 60 seconds. 

Note that the maximum allowed walking distances $d_r$ are related to the value of $\delta$. Specifically, $d_r$ values of 0.3 km, 0.4 km, and 0.5 km correspond to $\delta$ values of 300s, 360s, and 420s, respectively. %Figure \ref{fig:distribution} presents the distribution of the number of optional pickup locations to be considered within the maximum allowed walking distance for passenger $d_r$. The figure shows that 
Our experiments reveal that the average number of neighbor for each node is 10, 14, and 18, corresponding to $d_r$ values of 0.3 km, 0.4 km, and 0.5 km, respectively. 
% % The metric we use to compare the performance is the number of served requests, averaged over 3 runs for each method.
% The vehicle speed, $speed_v$ is 20km/h while the speed of the passenger $speed_r$ is 4km/h.
Note that the pickup and drop-off areas for some requests may overlap at the default speed. In such cases, we adjust the size of the areas to eliminate any overlap.

We compare our method, \METHODX, with state-of-the-art methods for solving RMP: NeurADP and HIVES. \METHODX\ allows flexible pickup and drop-off options, whereas NeurADP and HIVES assume fixed pickup and drop-off points.

%\begin{figure*}[htbp]
%    \centering
%    \includegraphics[scale=0.55]{Picture4.png}
%    \caption{Neighbour Number Distribution}
%    \label{fig:distribution}
%\end{figure*}

\subsection{Experimental results}
We first compare the number of served requests of \METHODX\ with those of NeurADP \cite{shah2020neural} and HIVES \cite{hao2022hierarchical}. We have averaged the results over three runs for each method. Table~\ref{table:1} provides the overall results with different settings of capacity, vehicle number and pickup delay. Here are the key observations:
\begin{enumerate}
    \item \METHODX\ consistently outperforms both NeurADP and HIVES across all settings. 
    %To provide perspective, even a 0.5\%-1\% is considered a significant improvement in city scale taxi on demand problems~\cite{xu2018large}. 
    Our average improvement over NeurADP and HIVES is 13\% and 6\% respectively, which is considered a significant improvement in city-scale taxi ride-pooling problems~\cite{xu2018large}. 
    \item When the capacity increases from 2 to 4, the percentage of improvement over NeruADP decreased from 16.75\% to 13.36\%. However, the improvement over HIVES increased from 5.42\% to 6.39\% over HIVES. This is because larger vehicle capacity provides more available resources, reducing the advantage over NeurADP, while in HIVES, the increase in capacity weakens the effect of clustering and neighboring, allowing \METHODX\ to benefit more from the increased flexibility.
    %More capacity means more available sources to serve the same amount of demand, so the advantage over NeurADP decreased. In HIVES, more available resources will weaken the effect of clustering and neighboring while in METHODX, more available resources will take more combinations into consideration, that is why METHODX outperformed HIVES when capacity increases.
    \item When the number of vehicles increases from 1000 to 2000, the improvement percentage over NeurADP decreased, while the improvement over HIVES increased. This is for the same reason as in the second observation: more available resources reduce the advantage over NeurADP but weaken the effect of clustering and neighboring in HIVES, allowing \METHODX\ to better utilize the additional vehicles.

    \item When the pickup delay increases from 300 to 420 seconds, the performance improvement over NeurADP increased from 13.36\% to 16.16\% , and the improvement over HIVES increased from 6.39\% to 7.31\%. That is because %the maximum allowed travel distance is positively related to pickup delay, a larger pickup delay makes the extended pickup and drop-off area larger, thus takes into more feasible pickup and drop-off locations into consideration, which can potentially improve the assignment policy.
    the maximum allowed travel distance is directly related to the pickup delay. A larger pickup delay increases the extended pickup and drop-off areas, allowing for more feasible options and improving the overall assignment strategy.
\end{enumerate}

The experimental results of average travel distance show similar behavior to the number of served requests, as illustrated in Table~\ref{table:3}. \METHODX\ significantly reduces the average travel distance compared to the baselines across all settings. This improvement comes from the flexibility provided by the extended pickup and drop-off areas,together with the route planning in \METHODX.

\subsubsection{Ablation Study}
In this part, we investigate the impact of extending the pickup and drop-off locations. We consider three variants: \textbf{1)} Extending the pickup area while keeping the drop-off at the original location (Pickup Only); \textbf{2)} Picking up at the original location while extending the drop-off area (Drop-off Only).
\textbf{3)} \METHODX\ with extensions for both the pickup and drop-off locations. 

Table~\ref{table:2} presents the results of the ablation study. As shown in the table, \METHODX\ with both extended pickup and drop-off location achieves best performance, and extending drop-off location brings greater improvements compared to only extending pickup location. This may be because the drop-off location affects the starting point for the assignment in the next time step, while the pickup location primarily influences the efficiency of completing currently assigned requests.

\section{Conclusion}
In this paper, we addressed a Ride-Pool Matching Problem (RMP) that incorporates flexible pickup and drop-off locations for passengers. We leveraged passengers' mobility by allowing them to walk to nearby locations to meet vehicles, while considering the long-term impacts of assignments and vehicle routes to achieve optimal matching outcomes.
% This flexibility not only enhances matching opportunities but also decreases vehicle detours.
% To further enhance the matching outcomes, we evaluated and identified the optimal matches and vehicle routes while considering their long-term impacts.

% By allowing passengers to move to nearby feasible locations, we optimized the assignment process for each vehicle, reducing unnecessary detours. Our method leverages dynamic assignment strategies that take into account both passenger mobility and vehicle constraints, such as capacity, pickup delays, and detour delays.

% Our method first generates all possible combinations between requests and vehicles. For each possible combination, we then compute an optimal route and optimal pickup and drop-off points for each passenger. These optimal routes will then passed to a value function to complete the assignment.

Our method first employed a tree-based approach to efficiently generate feasible combinations of requests for each vehicle. By incrementally creating combinations from smaller sizes to larger ones, infeasible matches are pruned at early stages to enhance efficiency.
% thus significantly reducing the search space and improving the efficiency.
Then, for each possible assignment, we applied regional route planning to determine the optimal routes for picking up and dropping off the assigned passengers. 
Finally, we utilized a reinforcement learning-based method to evaluate the matches and identify the optimal matches considering the long-term effects.
%The route planning ensures that vehicles respect capacity constraints and dynamically adjusts based on the assigned requests, leading to more flexible operations in city-scale scenarios.

Experiments on real-world city-scale datasets demonstrated that our approach
% , which combines efficient request assignment and route planning, 
outperforms existing leading methods by achieving up to a 13\% improvement in the total number of served requests and 21\% improvement in the average total distance. This result highlights the potential of incorporating passenger mobility and dynamic routing to enhance ride-pooling services in high-demand urban environments.

% In the real world, passengers may be reluctant to move to alternative locations without an appropriate incentive. We are considering a pricing model that provides a discount to passengers who are willing to move to optional pickup or drop-off locations. This incentive would encourage passengers to adjust their locations, enabling better assignment strategies. Although this might lower the fare for each individual request, it would allow vehicles to reduce travel time per trip and serve more passengers, ultimately increasing the overall revenue.

In the future, we plan to develop a pricing model for ride-pooling with flexible pickup and drop-off points. 
By offering discounts to passengers who walk to nearby pickup or drop-off locations, the pricing model has the potential to incentivize passengers to adjust their locations and lead to more optimized outcomes.

\bibliography{aaai25}

%\newpage
%\onecolumn
%\appendix
%
%\section{Structure of METHODX}

\end{document}